\documentclass{l4dc2024}

\newcommand{\R}{\mathbb{R}}
\renewcommand{\S}{\mathcal{S}}
\newcommand{\A}{\mathcal{A}}
\newcommand{\E}{\mathbb{E}}

\newcommand{\ReLU}{\small\text{ReLU}}

\title{Do No Harm: A Counterfactual Approach to Safe~Reinforcement~Learning}
\usepackage{times}
\usepackage{bbm}
\usepackage{multirow}
\usepackage{floatrow}
\newfloatcommand{capbtabbox}{table}[][\FBwidth]
\usepackage{booktabs}
\usepackage{caption}
\usepackage{blindtext}
\usepackage{lipsum}
\newtheorem{ass}[theorem]{Assumption}

\usepackage{enumitem}

\author{%
 \Name{Sean Vaskov} \Email{sean.vaskov@isee.ai}\\
 \addr ISEE AI, Cambridge, MA, 02139
 \AND
  \Name{Wilko Schwarting} \Email{wilko@isee.ai}\\
 \addr ISEE AI, Cambridge, MA, 02139
 \AND
 \Name{Chris L. Baker} \Email{chrisbaker@isee.ai}\\
 \addr ISEE AI, Cambridge, MA, 02139
}

\begin{document}

\maketitle

\begin{abstract}%
Reinforcement Learning (RL) for control has become increasingly popular due to its ability to learn rich feedback policies that take into account uncertainty and complex representations of the environment.
When considering safety constraints, constrained optimization approaches, where agents are penalized for constraint violations, are commonly used.
In such methods, if agents are initialized in, or must visit, states where constraint violation might be inevitable, it is unclear how much they should be penalized.
We address this challenge by formulating a constraint on the counterfactual harm of the learned policy compared to a default, safe policy.
In a philosophical sense this formulation only penalizes the learner for constraint violations that it caused; in a practical sense it maintains feasibility of the optimal control problem. 
We present simulation studies on a rover with uncertain road friction and a tractor-trailer parking environment that demonstrate our constraint formulation enables agents to learn safer policies than contemporary constrained RL methods.
\end{abstract}


\section{Introduction}
Learning-based control, particularly reinforcement learning (RL)~\citep{li2017deep,sutton2018reinforcement} has become increasingly popular due to its ability to learn powerful feedback policies that take uncertainty and complex environment representations into account.
However, learning policies for applications in which safe behavior is required remains a challenging problem.
Methods for safe RL typically fall into one of two categories: constrained optimization where the objective is to learn a policy whose output actions satisfy safety constraints; or shielding where a backup or safety-preserving policy corrects the agent's actions to be safe.
Constrained optimization techniques commonly use the method of Lagrange multipliers where the penalty for constraint violation is increased until the constraint is satisfied~\citep{altman1998constrained,chow2017risk,paternain2019constrained}.
Epigraph forms avoid numerical issues associated with Lagrange multipliers at the cost of introducing a state and estimator for a cost budget~\citep{so2023solving}.
In shielding approaches, a safety-preserving policy, often designed with control barrier functions~\citep{cheng2019end} or reachability analysis~\citep{shao2021reachability,selim2022safe,hsu2023sim} restricts the agent's actions to be safe during training.
In the context of safety, hierarchical methods~\citep{barto2003recent,lee2023adaptive} are related to shielding methods in that a higher-level policy learns when to execute a safety-preserving policy; however both policies can be trained with constrained optimization.

In simulated RL applications where constraint violation is not associated with tangible damage, optimization-based methods are preferred because they allow for more exploration as the agents are not limited to choosing the actions of the shielding policy.
A challenge associated with Lagrange methods is balancing the feasibility of the constraints with adjustment of the multipliers.
Enforcing that constraint violation never happens limits the domain policies can be used in and may be impossible in some environments; on the other hand, penalizing agents for constraint violations that are unavoidable will cause the multipliers to increase indefinitely and destabilize learning.

We address this challenge using counterfactuals~\citep{roese1997counterfactual,balke1994counterfactual}.
In the context of policy optimization, counterfactual metrics compare the outcome of one policy or action to another.
The conditional average treatment effect (CATE)~\citep{shpitser2012identification} compares the expected utility.
In RL, CATE-style objectives have been used for both credit~\citep{foerster2018counterfactual} and blame~\citep{triantafyllou2021blame} assignment.
Algorithms that minimize expected counterfactual regret have also been applied to extensive-form games with discrete action spaces \citep{zinkevich2007regret,brown2019deep}.
Counterfactual harm~\citep{richens2022counterfactual} compares the outcome of a selected action to a default action across all realizations of uncertainty, as opposed to in expectation, making it a more appropriate metric for safety than CATE.
\cite{beckers2022quantifying} offer a competing definition of harm where a default utility instead of action is used, and all actions are simultaneously compared, as opposed to the pairwise comparison of each action to the default.


In this paper we develop both CATE and harm-based constraints for safe RL.
We base our harm constraint off of~\cite{richens2022counterfactual}.
For dynamic systems, assuming a default policy or action is more appropriate than a default utility; in practice, the later would be computed based on analysis of the agent's state-action space and dynamics.
Additionally, we focus on single agent environments where we compare the learned policy to a default policy, so the pairwise comparison is complete.
Extensions to multi-agent environments and/or multiple policies may require incorporating elements of \citeauthor{beckers2022quantifying}'s definition.
To the best of our knowledge this is the first application of counterfactual reasoning, especially harm, to constrained RL.
Our contributions are three-fold:
\begin{itemize}[itemsep=0pt, topsep=3pt]
    \item Relating safe RL and viability theory to counterfactual reasoning techniques that analyze harm; this provides a philosophically and practically sound definition of safe behavior;
    \item A constraint based on the conditional average treatment effect (CATE), which we modify to reduce conservatism;
    \item A constraint restricting counterfactual harm, which addresses CATE's deficiency in safely handling uncertainty. In our formulation, we also connect notions of ``stability" in the counterfactual literature~\citep{lucas2015improved} to TD$(\lambda)$ methods~\citep{sutton2018reinforcement}.
\end{itemize}
The goal of this paper is to compare our proposed constraint formulations to those traditionally found in RL.
We focus on the Lagrange multiplier approach; investigating compatibility with and comparing to epigraph forms is left to future work.
We do not implement or compare to shielding and hierarchical methods; however, we discuss how to integrate our constraint formulations into them in Appendix \iffinalsubmission A.\footnote{An extended version, including the appendix, is available at \textcolor{red}{XXX}.} \else\ref{app:shield}. \fi
The paper is organized as follows. First, we give an overview of our notation, the constrained Markov decision process (CMDP) framework, and relevant concepts from causal reasoning and viability theory.
Next, we propose three alternative constraint formulations for safe RL, the last two of which are contributions of this paper.
Finally, we present two simulation studies showcasing the effectiveness of our constraints.

\section{Preliminaries}
In this section, we first define our notation and an instance of the general class of problems this paper solves. We then introduce concepts from causal modeling and viability theory that are relevant to our formulation and results.
A Markov decision process~(MDP) is a tuple $(\S,\A,p_s,r,\gamma,p_0)$ where $\S$ and $\A$ are compact state and action spaces, $p_s(\cdot|s_t,a_t)$ is the transition probability distribution conditioned on a state and action at time $t$, $p_0$ is the initial state distribution, $r:\S\times \A \rightarrow \R$ is an instantaneous reward function, and $\gamma\in(0,1)$ is a discount factor.
In the safe RL setting, we introduce a set of constraint functions $g_i:\S\rightarrow \R, i\in\{1,...,n_c\}$ whose 0-sublevel sets represent safe states.
A policy is a mapping from states to actions and can either be deterministic or stochastic.
Trajectories generated by the state transition distribution under a policy, $\pi$, are written as $s_{0:T}^\pi$.
We can learn a safe, reward-maximizing policy by solving the following optimal control problem:
\begin{subequations}\label{eq:ocp}
\begin{align}
    \pi^*=\underset{\pi}{\text{argmax}}\ & \E\left[\sum_{t \geq 0}\gamma^t r(s_t,a_t)\right]\\
    &s_0\sim p_0,\ s_{t+1}\sim p_s(\cdot|s_t,a_t),\ a_t\sim \pi(\cdot|s_t)
    \\\label{eq:ocp_constraints}
   \text{s.t. }&
 \E\left[\max_{t\geq 0} \gamma^t g_i(s_t)\right]\leq 0,\ i \in\{1,...,n_c\}.
\end{align}
\end{subequations}
\begin{remark}
    Program~\eqref{eq:ocp} differs from the classical Constrained Markov Decision Process~\citep{altman1998constrained}, which uses cumulative constraints (\textit{i.e.} $\E[\sum_{t \geq 0}\gamma^t g_i(s_t)]\leq 0$). Here,
    \eqref{eq:ocp_constraints} ensures each constraint is non-positive at all times---the standard formulation for constrained optimal control problems~\citep{johansen2011introduction}.
    All proposed constraint formulations in this paper can use either the max or sum operator.
\end{remark}

Since \eqref{eq:ocp} will be solved with RL we refer to $\pi$ as the \emph{learner policy}.
Throughout this paper, we default to using the expectation operator without subscripts to indicate sampling $\pi$, with the state transition distribution $p_s$; when needed we use subscripts to provide details, \emph{e.g.} we write $\E_{p_0,\mu}$ to indicate that we are sampling actions from policy $\mu$ and initial states from $p_0$.
To solve \eqref{eq:ocp} we leverage the method of Lagrange multipliers, and iteratively solve the unconstrained problem~\citep{paternain2019constrained}:
\begin{subequations}
\label{eq:lagrange_ocp}
   \begin{align}
(w^*,\pi^*)=\text{arg }\underset{w\in \R_+^{n_c}}{\min}\ \underset{\pi}{\max}\ & \E\left[\sum_{t \geq 0}\gamma^tr(s_t,a_t)\right]-\sum_{i=1}^{n_c}w_i\E\left[\max_{t \geq 0} \gamma^t g_i(s_t)\right]\\
   \text{s.t. }&
   s_0\sim p_0,\ s_{t+1}\sim p_s(\cdot|s_t,a_t),\ a_t\sim \pi(\cdot|s_t).
\end{align} 
\end{subequations}
Since the penalty for constraint violation, $w$, is automatically adjusted, the Lagrangian method is preferred to hand tuning in applications where constraint satisfaction conflicts with the performance objective~\citep{roy2021direct}.
For notational brevity, we hereafter refer to a single constraint, $g$, as one can simply repeat our formulations for multiple constraints.

We next introduce assumptions and terminology related to counterfactual reasoning, structural causal models, and viability theory.
\begin{ass}\label{ass:scm}
Let $(\Xi,\mathcal{F},P)$ be a probability space.
The state transition distribution can be modeled by a deterministic function, $f:\mathcal{S}\times \mathcal{A}\times \Xi\rightarrow\mathcal{S}$, with exogenous noise variables,
\begin{align}\label{eq:scm_model}s_{t+1}=f(s_t,a_t,\xi_t),\ \xi_t\in \Xi.
\end{align}
This assumption is always possible to satisfy using auto-regressive uniformization~\citep[Lemma 2]{buesing2018woulda}.
Partially observable environments and the evolution of stochastic policies can be modeled by making the same assumption about the observation and action distributions.
\end{ass}
Assumption~\ref{ass:scm} allows us represent the CMDP as a \emph{structural causal model}~\citep{pearl2000models} where the state transitions, constraints, and rewards are endogenous variables and the noise values are exogenous variables.
We refer to a trajectory of exogenous variables, $\xi_{0:T}$, as a \emph{realization of uncertainty}.
A \emph{counterfactual query} is a triple of an observation, intervention, and outcome $(X,I,Y)$, that asks the question: given observation 
$X$, what would $Y$ have been under intervention $I$?
The counterfactual query can be answered with \emph{counterfactual inference}~\citep{balke1994counterfactual}.
Counterfactual inference has three steps: infer the distribution of the (potentially unobserved) exogenous variables, $p_\xi(\cdot|X)$; simulate the system using \eqref{eq:scm_model} with the inferred distribution and intervention; and compute the outcome, $Y$.
In our context $X$ will be trajectories $s_{0:N}^\pi$, $I$ will be a safe policy, and $Y$ will be the constraint values.

In RL applications the practitioner often chooses the initial state distribution, $p_0$.
To understand how this relates to the feasibility of \eqref{eq:ocp} we introduce concepts from viability theory~\citep{aubin2011viability}.
Let $p_\xi$ be a probability distribution that describes realizations of uncertainty.
The \emph{viability kernel} is the set of initial states from which the constraints can be satisfied:
\begin{align}\label{eq:viab}
    \text{Viab}_{p_\xi}= \{s_0\in \S ~|~ \forall\ \xi_{0:\infty}\sim p_\xi,\ \exists \{a_t\}_{t\geq 0}  \small\text{ s.t. } \max_{t\geq 0}g(s_t)\leq 0, \small\text{ where }s_{t+1}=f(s_t,a_t,\xi_t)\},
\end{align}
We define the viability kernel for a policy, $\mu$, as
\begin{align}\label{eq:viab_mu}
    \text{Viab}_{p_\xi}^\mu= \{s_0\in \S ~|~ \forall\  \xi_{0:\infty}\sim p_\xi,\ \max_{t\geq 0}g(s_t)\leq 0,\small\text{ where }s_{t+1}=f(s_t,\mu(s_t),\xi_t)\},
\end{align}
Counterfactual inference allows us to understand how an agent should be penalized for constraint violation.
If the agent was in the viability kernel, there is an intervention on the agent's actions that can satisfy the constraint so the agent should be penalized harshly.
As the agent leaves the viability kernel, the change in outcome for any intervention diminishes, so the agent's penalty should as well.
In the following section we present reformulations of \eqref{eq:ocp_constraints} that capture this relationship.
\section{Formulations}\label{sec:formulation}
In this section, we outline three alternative formulations of the constraint~\eqref{eq:ocp_constraints}, the last of which is the main contribution of this paper.
We are particularly interested in the relationship between the initial state distribution, $p_0$, the viability kernel~\eqref{eq:viab}, and the values of the constraints~\eqref{eq:ocp_constraints}.
Leaving~\eqref{eq:ocp_constraints} as is requires us to engineer the initial state distribution to be a subset of the viability kernel in order for~\eqref{eq:ocp} to be feasible.
In this case, agents may never or rarely visit certain boundaries of the viability kernel, and produce unsafe behavior if they do.
Furthermore, in some environments, engineering $p_0$ to be feasible may be impossible if uncontrollable forces or bad actors cause the agent to exit the viability kernel (regardless of how it is initialized).
We reformulate~\eqref{eq:ocp_constraints} to allow agents to be initialized in and out of the viability kernel.
In order to do so, we make the following assumption:
\begin{ass}\label{ass:mu}
    We are given access to a default action or policy, $\mu$, that the agent can execute to avoid constraint violation.
    In the event that agents have exited the viability kernel, the default policy can minimize the severity of violation; for example, a vehicle could brake to reduce the impact energy of a collision.
    We assume there is one default $\mu$ for all constraints.
    $\mu$ can be hand-designed, computed using reachability analysis~\citep{bansal2017hamilton,fisac2019bridging}, learned with behavior cloning~\citep{torabi2018behavioral}, or a combination thereof.
\end{ass}

Our first reformulation of \eqref{eq:ocp_constraints} makes the threshold a function of the default policy:
\begin{align}\label{eq:initial_cons}
\begin{split}
 \E_{p_0,\pi}\left[\max_{t \geq 0} \gamma^t g(s_t)\right]\leq
 \E_{p_o,\mu}\left[\ReLU\left( \max_{t \geq 0} \gamma^tg(s_t)\right)\right].
\end{split}
\end{align}
This option ensures the feasibility of~\eqref{eq:ocp} but will not result in safe policies because it only enforces the learner policy to violate the constraints less than the default policy in expectation over $p_0$ and $p_s$.
Using a risk metric such as CVaR$_\alpha$~\citep{chow2017risk} will not address this problem since we can apply the same argument about averaging to violations occurring in the $\alpha$-quantile.
In the following subsections we propose constraints that remedy the effects of averaging over $p_0$ and $p_s$.

\subsection{Clipped Conditional Average Treatment Effect}
The second option is a modified version of 
the conditional average treatment effect~(CATE)~\citep{shpitser2012identification}, which enforces the learner policy to be safer than the default policy in expectation:
\begin{align}\label{eq:ccate}
\text{CCATE}_g^\mu(s_t)=&\mathbb{E}_\pi\left[\max_{\tau>t}\gamma^{\tau-t} g(s_\tau)\right]-\text{ReLU}\left(\mathbb{E}_\mu\left[\max_{\tau >t}\gamma^{\tau-t} g( s_\tau)\right]\right).\end{align}
We refer to this as the \emph{clipped conditional average treatment effect}~(CCATE).
The clipping ensures that the learner policy is not required to satisfy the constraints more than the default policy, only to violate them less, which reduces conservatism while preserving safety.
We replace \eqref{eq:ocp_constraints} with $\mathbb{E}[\max_{t\geq 0}\gamma^{t}\text{CCATE}_{g}^\mu(s_t)]\leq 0$.
CCATE is stronger than \eqref{eq:initial_cons} since it is enforced at each state the agent visits, instead of averaging over $p_0$.
CCATE can be implemented efficiently if we precompute a function approximation of $\mu$'s expected constraint violation.
However, CCATE only compares the violations in expectation over $p_s$---it is not robust to dynamics uncertainty.


\subsection{Counterfactual Harm}\label{sec:formulation_harm}
The safest option uses \emph{counterfactual harm}~\citep{richens2022counterfactual}, which we apply as a constraint as follows: we first define the counterfactual harm at a given state, $s_t$, as
\begin{align}\label{eq:harm}H_g^\mu(s_t)=&\mathbb{E}\left[\small\text{ReLU}\!\left(\max_{\tau>t}\gamma^{\tau-t} g(s_\tau)-\small\text{ReLU}\!\left(\mathbb{E}_{\substack{ \tilde s_{\tau+1} =f(\tilde s_\tau,\mu(\tilde s_\tau), \tilde \xi_\tau)\\\tilde\xi_\tau \sim p_\xi(\cdot |s_{t+1},s_t)\\ \tilde s_t=s_t}}\left[\max_{\tau>t}\gamma^{\tau-t} g(\tilde s_\tau)\right]\!\right)\!\right)\right];\end{align}
then we replace \eqref{eq:ocp_constraints} with the equality constraint $\mathbb{E}[\max_{t\geq0}\gamma^{t}H_{g}^\mu(s_t)]=0$.
\begin{remark}\label{rem:mono_harm}
    Composing any monotonic function satisfying $f(0)\!=\!0$, with counterfactual harm does not change the underlying constraint. E.g., the indicator function constrains the probability of harm.
\end{remark}
Our definition of harm differs from~\citeauthor{richens2022counterfactual}'s in two ways.
First, as in~\eqref{eq:ccate}, we clip the outcome of $\mu$ for the constrained setting. Second, we compare outcomes over an infinite time horizon; we require this when exiting the viability kernel takes multiple timesteps.
Counterfactual harm differs from CCATE in two ways.
First, the exogenous variables are conditioned on the observed trajectory ($p_\xi(\cdot |s_{t+1},s_t)$ in~\ref{eq:harm}); this is important because we want to compare outcomes had the same or a similar scenario unfolded.
Second, there is an additional clipping inside the expectation, which enforces the constraint for all realizations of uncertainty.
Defining $p_\xi$ is an active area of research.
If the exogenous variables are observable, we can use the twin-network model~\citep{pearl2000models}, where their values are equivalent, $p_\xi(\cdot |s_{t+1},s_t)\leftarrow\delta_{\xi_t}$.
\cite{lucas2015improved} propose a stability parameter that probabilistically interpolates between sampling the observed or random noise.

To estimate~\eqref{eq:harm} we perform counterfactual inference for $N$ steps, summarized in Algorithm~\ref{alg:cf_estimate}.
We estimate the infinite time horizon outcome by using a learned approximation of the default policy's \emph{constraint value function}, $V_g^\mu(s_t)\approx \mathbb{E}_\mu [\max_{\tau \geq t} \gamma^{\tau-t} g(s_\tau)]$, and the recursive estimator
\begin{align}\label{eq:tdl_max}
\hat{V}_{g}^{\mu,\lambda}(\tilde s_\tau)=\max\{g(\tilde s_\tau),\gamma(\lambda \hat{V}_{g}^{\mu,\lambda}(\tilde s_{\tau+1})+(1-\lambda)V_{g}^\mu(\tilde s_{\tau+1}))\},
\end{align}
with $\hat{V}_{g}^{\mu,\lambda}(\tilde s_{\tau+N})=V_g^\mu(\tilde s_{\tau + N})$ and $\lambda\in[0,1)$.
In RL, \eqref{eq:tdl_max} is referred as a TD($\lambda$) method~\citep{sutton2018reinforcement}, but its application to the max operator is a contribution of this paper.
Previous works using the max operator in RL~\citep{fisac2019bridging,li2022infinite} only present single step $(\lambda=0)$ estimates.
$\lambda$ is analogous to the stability parameter in~\cite{lucas2015improved} since increasing (decreasing) it weights trajectories generated from the observed (marginal) exogenous variables higher.
We also apply~\eqref{eq:tdl_max} to estimate the learner policy's constraint violation and the maximum harm over the episode as follows:
\begin{align}\label{eq:harm_return}
   \hat V_{H}^{\lambda} (s_t)=\max\{ \ReLU(\hat V_g^{\pi,\lambda}(s_t)-\ReLU(\hat V_g^{\mu,\lambda}(s_t))),\gamma (\lambda \hat V_{H}^{\lambda}(s_{t+1})-(1-\lambda)V_{H}(s_{t+1}))\}.
\end{align}


\begin{algorithm2e}[!htb]
\small
\caption{$N$-step Counterfactual Inference}\label{alg:cf_estimate}
\DontPrintSemicolon
\LinesNumbered
\KwData{State trajectory $s_{t:t+N}^\pi$, default policy $\mu$, constraint $g$, value function $V_g^\mu$, stability parameter $\lambda$, $N$}
 \KwResult{Counterfactual outcome: $\max_{\tau \geq t} \gamma^{\tau -t}g(s_t)$, of intervention: replace $\pi$ with $\mu$
}
$\tilde s\gets \{s_{t}\}$\;
\For {$\tau\in \{0,...,N-1\}$}{
$\tilde s_{\tau+1}=f(\tilde s_\tau,\mu(\tilde s_\tau),\tilde \xi_\tau),\text{ with } \tilde \xi_\tau \sim p_\xi(\cdot|s_{\tau+t+1},s_{\tau+t})$\tcp*[f]{$p_\xi\gets \delta_{\xi_t}$ if $\xi$ observable}\;
$\tilde s \gets \tilde s \cup \tilde s_{\tau+1}$
}
\Return $\hat V_{g}^{\mu,\lambda}(s_t)$; estimated with $\tilde s$, \eqref{eq:tdl_max}, and $\hat V_{g}^{\mu,\lambda}(\tilde s_{t+N})=V_g^\mu(\tilde s_{t+N})$\;
\end{algorithm2e}
\section{Learning Implementation}
We solve both CCATE and Harm constrained algorithms using reinforcement learning and the Lagrange multiplier technique \eqref{eq:lagrange_ocp}.
For the policy optimization step in Algorithm \ref{alg:ccrl}, we use PPO~\citep{schulman2017proximal} with the addition of counterfactual inference.
We use separate critics for the reward, default and learner constraints, and counterfactual harm (or CCATE).
\begin{algorithm2e}
\small
\caption{Counterfactual Harm Constrained Policy Update (Actor Critic)}\label{alg:ccrl}
\DontPrintSemicolon
\LinesNumbered
\KwData{$r$, $g$, $\lambda$, $M$, $N$, Lagrange multipliers $w$, default policy $\mu$, critics $V_r,\ V_g^\pi,\ V_g^\mu,\ V_{H}$ and actor $\pi$}
\KwResult{Optimized policy $\pi$}
\For {$k\in \{0,...,M\}$}{
Step forward environment with $a_{t+k}\sim \pi_\text{old}(s_{t+k})$ and  store $s_{t+k}$, $a_{t+k}$, $r(s_{t+k},a_{t+k})$, $g(s_{t+k})$\;
}\
Estimate reward advantage $\hat A_r^{\lambda}$ with GAE and constraint violations, $\hat V_g^{\pi,\lambda}$ with \eqref{eq:tdl_max}\;
\For {$k\in \{M,...,0\}$}{
Estimate $N$-step counterfactual outcome, $\hat V_g^{\mu,\lambda}(s_{t+k})$, with Algorithm \ref{alg:cf_estimate} \label{alg:ccrl_inference}\;
Estimate maximum harm over episode $\hat V_{H}^{\lambda}(s_{t+k})$ with \eqref{eq:harm_return}\label{alg:ccrl_harm_return}\;
}
Optimize $\pi$ with \eqref{eq:surrogate_loss} plus critics with L2 (or other suitable) loss\label{alg:ccrl_opt}, e.g. ${V^{\pi}_g}$ with $\sum_i (\hat V_g^{\pi,\lambda}(s_i)-V_g^\pi(s_i))^2$\label{alg:ccrl_loss}\; 
\end{algorithm2e}
After collecting data from the learner rollouts
we estimate returns of the reward using generalized advantage estimation~(GAE)~\citep{schulman2015high} and constraint violation using \eqref{eq:tdl_max}.
We apply Algorithm \ref{alg:cf_estimate} at each state the agent visits (Line \ref{alg:ccrl_inference}), and compute the counterfactual harm with~\eqref{eq:harm_return}.
Alternatively, per Remark \ref{rem:mono_harm}, the $\gamma$-discounted probability of harm can be optimized by applying an indicator function to the computed harm. 
For CCATE we replace $\hat V_g^{\mu,\lambda}$ with $V_g^\mu$ and remove the outer ReLU.
To optimize the actor, $\pi$, we use the loss (omitting clipping terms in PPO for brevity):
\begin{align}\label{eq:surrogate_loss}
    \mathcal{L}_\text{actor}=\frac{-1}{N_\text{batch}}\sum_i\left(\frac{\pi(a|s_i)}{\pi_\text{old}(a_i|s_i)}( \hat A^{\pi,\lambda}_r(s_i)-w\cdot (\hat V_H^{\lambda}(s_i)-V_H(s_i)))+\mathcal{E}(\pi(a|s_i))\right),
\end{align}
which maximizes the advantage of the unconstrained objective \eqref{eq:lagrange_ocp}.
$\mathcal{E}(\pi(a|s_i))$ in \eqref{eq:surrogate_loss} is a small reward for entropy of the policy.
The critics are trained by minimizing the mean squared error or cross-entropy loss (in chance-constrained settings) from returns computed with \eqref{eq:tdl_max} (Line \ref{alg:ccrl_loss}). Convergence proofs for using \eqref{eq:tdl_max} to estimate the constraint value function are given in 
Appendix \iffinalsubmission B.\else\ref{app:max_proof}.\fi

\section{Experiments}\label{sec:experiments}
In this section we compare the performance of different constraint formulations.
We compare to two formulations from the literature: Direct Behavior Specification, DBS~\citep{roy2021direct}, which composes an indicator function with the constraint; and Instantaneous Constrained RL, IC~\citep{li2021augmented}, which applies a clipping function.
These methods use cumulative constraints and initialize agents within a heuristically chosen subset of the viability kernel referred to as the \emph{feasible initialization}, which is common practice in RL.
The additional formulations are as follows: MC\_0 uses \eqref{eq:ocp_constraints} and the feasible initialization; MC uses \eqref{eq:initial_cons}; CCATE \eqref{eq:ccate}; and HARM \eqref{eq:harm}.
We also test chance-constrained formulations of \eqref{eq:ocp_constraints}, CC\_0, and \eqref{eq:initial_cons}, CC, which apply an indicator function to the constraint; and  CCATE\_C and HARM\_C which follow from Remark \ref{rem:mono_harm}.

For the experiments we assume we have access to the simulation model, meaning \eqref{eq:scm_model} is the environment model and the exogenous variables are observable.
This paradigm is applicable to RL methods that use an accessible simulation model to generate synthetic data~\citep{buesing2018woulda,levine2020offline}.
The goal of this study is to understand the benefits of the counterfactual harm constraint, not to produce the best possible agents for benchmarking, so we use hand-designed default policies and quantify their suboptimality.
All methods use the Lagrange multiplier approach \eqref{eq:lagrange_ocp} with PPO.
To improve efficiency we increase the multiplier update rate as training progresses; analogous to the penalty parameter in~\cite{li2021augmented}.
We train all algorithms using the same hyperparameters; which we tuned to achieve efficient training as a group---we did not optimize hyperparameters for methods individually.
If the performance worsened at the end of training, we select the checkpoint with the least constraint violations followed by the highest success rate.

We compare the methods on two environments, a rover navigating a U-shape track with uncertain road friction and a tractor-trailer parking task.
We compare the ability of the learned policies to stay within the viability kernel, the severity of constraint violation, and the success rate.
Analytically computing the true viability kernel~\eqref{eq:viab} is intractable, so we use the \emph{default viability kernel}, Viab$_{\delta}^\mu$~\eqref{eq:viab_mu}, where the noise distribution follows the twin network model~\citep{pearl2000models}.
We initialize 20,000 agents randomly in and out of Viab$_{\delta}^\mu$ and compare the recall and discovery rate (DR) of the learner policy.
The recall assesses safety by counting instances where the learner policy violates the constraints and the default policy does not.
DR counts instances where the learner policy satisfies the constraints and the default policy does not; which quantifies how much Viab$_{\delta}^\mu$ under-approximates the true viability kernel.
We also measure the probability that agents reach their goal, ``Success", safely given they start within Viab$_{\delta}^\mu$; and the probability that they incur harm, \textit{P}$_\text{harm}$.
We round all values to the nearest percent.
For the tractor trailer environment, we plot the cumulative distributions of harm and constraint violation.
Extended descriptions of each environment, plots of performance and safety during training, and further discussion are in Appendix \iffinalsubmission C.\else\ref{app:results}.\fi
\subsection{Rover}
\begin{figure}[!htb]
\begin{floatrow}
\capbtabbox{%
\small
\begin{tabular}{|l|c|c|c|c|c|}
\hline
\multicolumn{1}{|c|}{} &  Rec&  DR        & Success & \multicolumn{1}{l|}{\textit{P}$_\text{harm}$} \\ \hline
DBS       &             0.96 & 0.02     &            0.95            & 0.06                          \\ \hline
IC             &   0.93        & 0.01       & 0.93            & 0.09                          \\ \hline
MC\_0                &      0.93   & 0.02          & 0.92            & 0.10                          \\ \hline
CC\_0              &      0.96     & 0.02          & 0.96            & 0.05                          \\ \hline
MC              &       0.96       &     \textbf{0.03}       & 0.96            & 0.06                          \\ \hline
CC               & 0.97        & \textbf{0.03}              & 0.97            & 0.04                          \\ \hline
CCATE                &      0.97   & 0.02          & 0.97           & 0.05                          \\ \hline
CCATE\_C                &      0.98   & \textbf{0.03}          & 0.89            & 0.03                          \\ \hline
HARM                 & 0.96   &0.02     & 0.96   & 0.05             \\ \hline
HARM\_C                 & \textbf{0.99}   &\textbf{0.03}      & \textbf{0.99}   & \textbf{0.02}                 \\ \hline
\end{tabular}
}{%
  \caption{Viability Statistics for rover}
  \label{tab:rover_stats}
}
\ffigbox{%
    \includegraphics[width=0.45\textwidth]{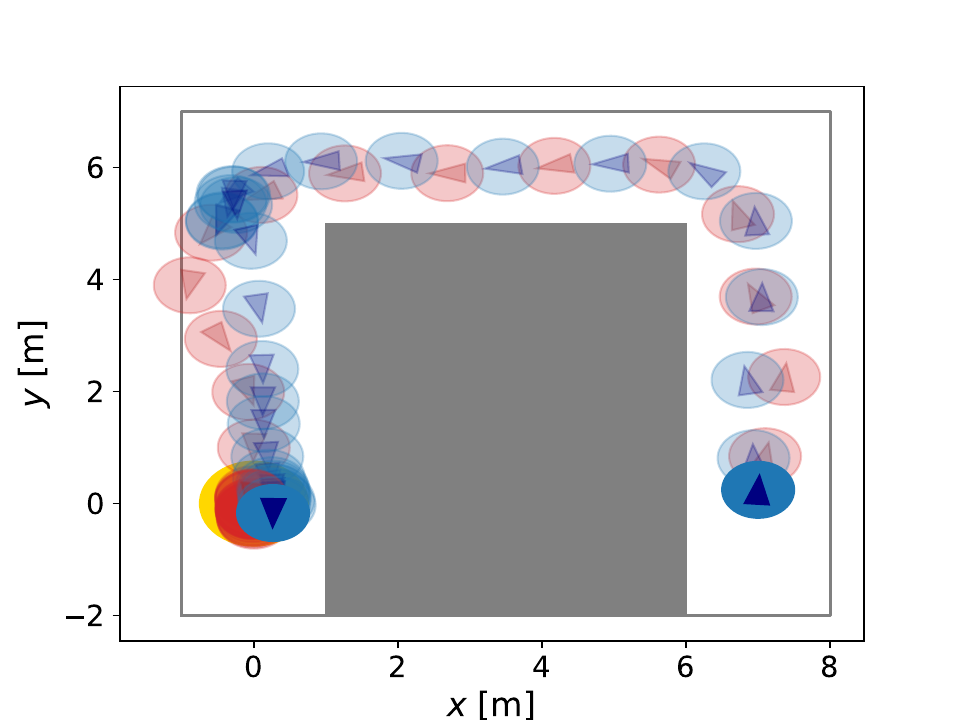}
}{%
  \caption{CC (red) and HARM\_C (blue) policies}
  \label{fig:rover_trajectory}
}
\end{floatrow}
\end{figure}

In this environment a rover is tasked to navigate a U-shaped corridor and reach a goal depicted in Figure \ref{fig:rover_trajectory}.
The rover is modeled as a kinematic bicycle model with a circular footprint.
The action commands are longitudinal acceleration and wheel angle; which, along with the speed, are limited in the dynamics model.
The surface has a friction coefficient randomly sampled from the interval $[0.3,1.0]$ for each environment.
The longitudinal and lateral acceleration are clipped to lie within the friction circle.
The rover receives a noisy observation of its state and the friction coefficient.
Noise is also added to the rover's action commands and the true friction coefficient at each timestep.
The rover reaches the goal if it arrives within 0.5 m of the origin with a speed $\leq$ 0.1 m/s.
The constraint is the signed distance of the center of mass to the outer boundary and obstacle in the middle.

The feasible initialization randomly places agents in the center of the corridor at 0 speed and a heading within $\pm \pi/2$ relative to the corridor direction.
To initialize agents out of the viability kernel, we allow the center of mass to be anywhere in the corridor and for the agents to have nonzero speed; as a result 50\% of agents are initialized out of Viab$_{\delta}^\mu$.
The policy is a normal distribution with a state-independent variance parameter for each action.
The network architecture used for the critics and the policy mean has 2 layers of dimension 256 and Tanh activation functions.
The default policy is braking with maximum deceleration and steering toward the corridor center. We use a time discretization of 0.5 s and episode length of 100 steps. We train with 3,000 parallel environments for 15,000 policy updates and use $N=5$ counterfactual steps and $M=24$ steps for each rollout in Algorithm \ref{alg:ccrl}.
The addition of counterfactual inference and critics increases training time by 5\%.
The results are shown in Table \ref{tab:rover_stats}.
\thisfloatsetup{subfloatrowsep=none}

\subsection{Tractor-Trailer Parking}
\begin{figure}[!htb]
\begin{floatrow}
\capbtabbox{%
\small
\begin{tabular}{|l|c|c|c|c|c|}
\hline
\multicolumn{1}{|c|}{} &  Rec&  DR        & Success & \multicolumn{1}{l|}{\textit{P}$_\text{harm}$} \\ \hline
DBS       &             0.93 & 0.07     &           0.90           & 0.18                          \\ \hline
IC             &   0.90        & 0.06       & 0.89            & 0.20                          \\ \hline
MC\_0                &      0.91   & 0.07          & 0.90            & 0.18                          \\ \hline
CC\_0              &      \textbf{0.94}     & 0.07          & 0.91            & 0.16                          \\ \hline
MC              &       0.80       &     \textbf{0.09}       & 0.79            & 0.23                          \\ \hline
CC               & 0.93        & 0.08              & 0.92            & 0.19                          \\ \hline
CCATE                &      0.92   & 0.08         & 0.90            & 0.14                          \\ \hline
CCATE\_C                &   0.93   &    0.08      &    0.92    &           0.18              \\ \hline
HARM              & \textbf{0.94}    & \textbf{0.09}        & \textbf{0.93}   & \textbf{0.11}                 \\ \hline
HARM\_C     & 0.93        &  0.08  &    \textbf{0.93}        &     0.17            \\ \hline
\end{tabular}
}{%
  \caption{Viability statistics for tractor-trailer}
  \label{tab:tractor_trailer_stats}
}
\ffigbox{%
\includegraphics[width=0.45\textwidth]{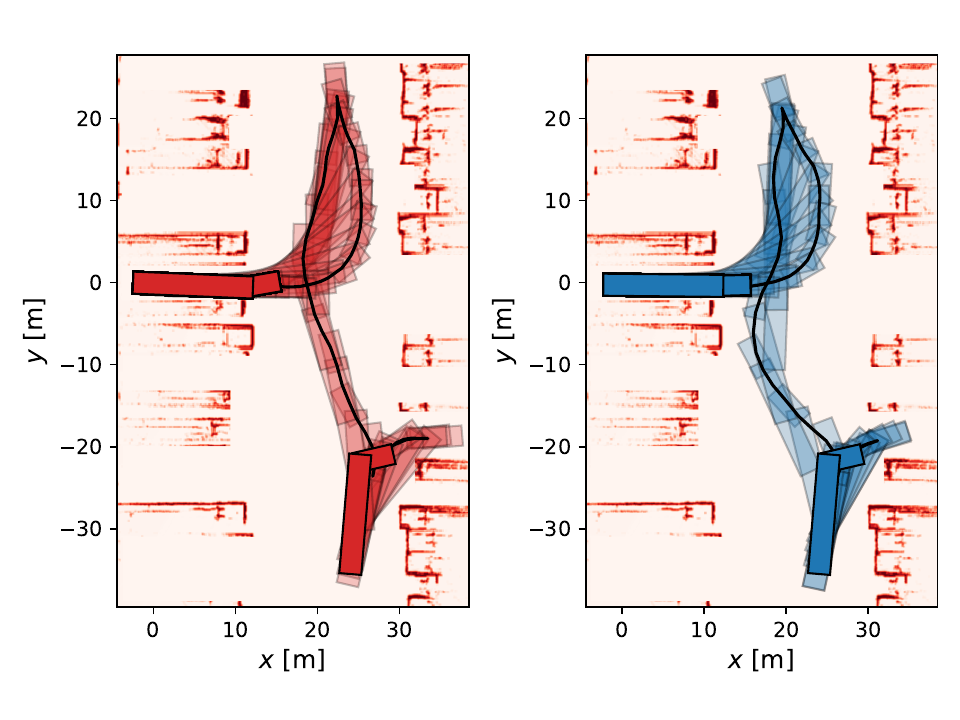}
}{%
  \caption{CC\_0 (red) and HARM (blue) policies}
  \label{fig:tractor_trailer_trajectory}
}
\end{floatrow}
\end{figure}

The second environment is a tractor-trailer parking task where the ego agent must park a trailer into a variety of spots as shown in Figure \ref{fig:tractor_trailer_trajectory}.
The actions are the acceleration and wheel angle rate and are limited, along with the speed, within the dynamics.
Normally distributed noise proportional to the speed is added to the dynamics.
Additionally the trailers take random, unobservable, wheelbases.
The static environment is an occupancy grid and the agent performs raycasting to return the distance to the closest occupied points and their relative velocity.
Noise proportional to their distance is added to the observation.
The constraint we consider is a modified version of the signed distance to obstacles where, when violated, we multiply it by 1+$v^2/2$.
The addition of $v^2/2$ penalizes collision severity by assuming the tractor absorbs 50\% of its kinetic energy upon impact.

\begin{figure}[!htb]
         \includegraphics[width=0.45\textwidth]{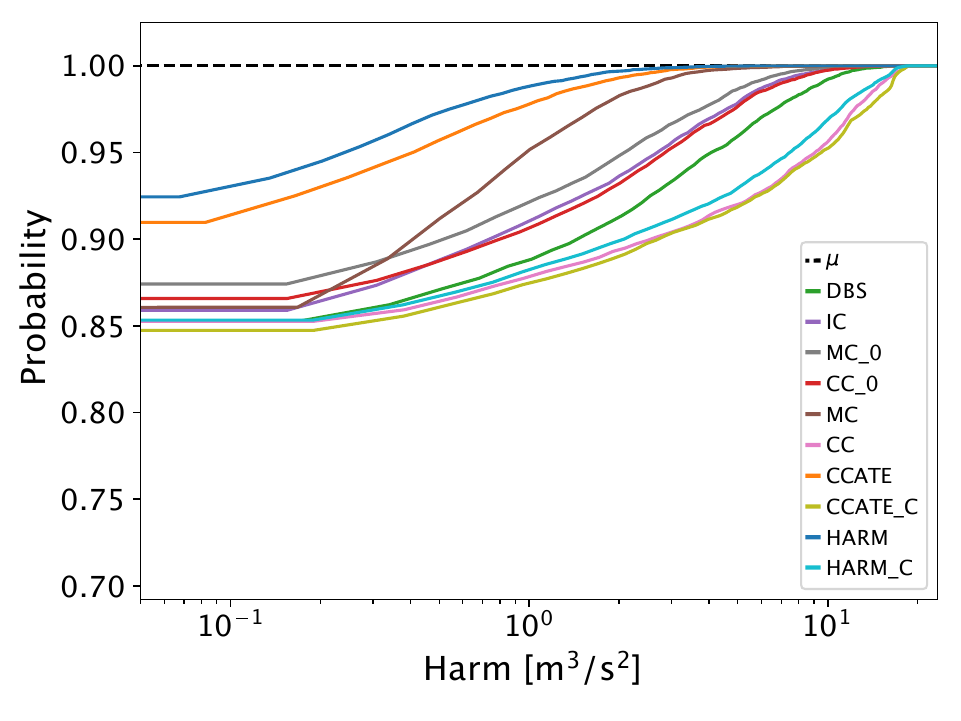}
\qquad
         \includegraphics[width=0.45\textwidth]{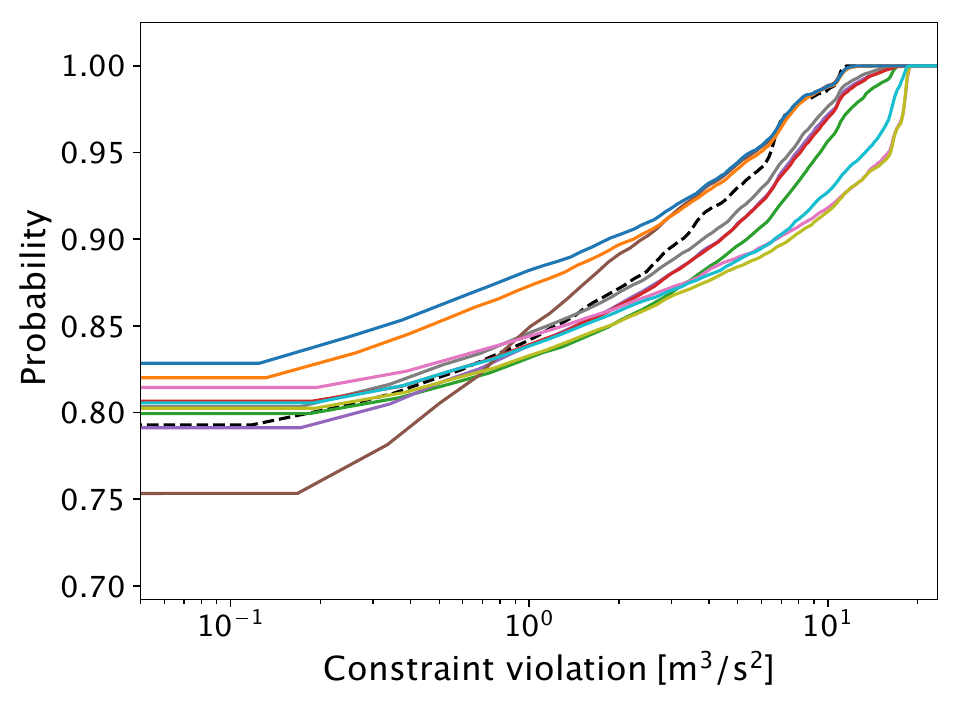}
\caption{\label{fig:cdfs}Cumulative distribution of harm (left) and constraint violations (right) for tractor-trailer. The black dashed lines are generated by executing the default policy, $\mu$, from the initial states.}
\end{figure}

Agents are initialized in collision-free states near their parking spots with random positions, orientations, velocities and wheel angles.
22\% of the agents are initialized outside of the default viability kernel.
The feasible initialization sets their velocity to 0. The default policy, $\mu$, is braking with maximum deceleration and maintaining the current wheel angle.
The network architecture consists of a shared two-layer encoder network that processes the lidar and state observations, then passes the concatenated output to the actor and critic networks, which are similar to those used for the rover.
The training hyperparameters are the same as the rover except the episode length is extended to 300 steps and the number of counterfactual steps is $N=4$.
The agents are trained with 10,000 parallel environments for 15,000 policy updates.
The addition of counterfactual inference and critics increases training time by 25\%.
The performance statistics are shown in Table \ref{tab:tractor_trailer_stats} and the cumulative distributions of the harm and constraint violation in Figure \ref{fig:cdfs}.\\
\thisfloatsetup{subfloatrowsep=none}
\subsection{Discussion}\label{sec:discussion}
HARM\_C and HARM achieve the highest recall, success rate, and lowest \textit{P}$_\text{harm}$ in the rover and tractor-trailer environments, respectively.
The cumulative distribution of HARM in the tractor-trailer environment upperbounds those of the other methods, indicating it incurs less constraint violations both in terms probability and severity.
Optimizing the probability performs worse in the tractor-trailer environment because the impact-energy constraint requires the agent to exactly apply the maximum deceleration to achieve 0 harm, which PPO cannot easily sample from a normally distributed policy.
It is easier to achieve 0 harm with the signed distance constraint of the rover, by sampling actions that steer the agent toward the centerline. 

In the tractor-trailer environment, CC\_0 matches and DBS is close to achieving the same recall, despite using the feasible initialization; however, they incur severe constraint violations as shown in Figure \ref{fig:cdfs}.
MC and CC are able to satisify their constraints but do not generate safe policies due to the issues with \eqref{eq:initial_cons} outlined in Section \ref{sec:formulation}.
Compared to those for the rover, every method's DR is higher due to the higher speed and suboptimality of the default policy.
Considering a default policy with evasive steering action would reduce the DR, although conditioning on the default policy did not impede the safety of HARM---in this experiment, the DRs are similar across methods and HARM incurred the fewest and least severe constraint violations. 
HARM and/or HARM\_C also achieves the highest success rate in both environments.
We hypothesize that the counterfactual constraint enables the policy to learn a better understanding of the viability kernel since it is only penalized for state-action sequences that cause or increase constraint violation.
\section{Conclusion}
We presented novel counterfactual constraints for safe reinforcement learning; most notably, a constraint on counterfactual harm.
Our formulation performs counterfactual inference to only penalize agents when they cause a constraint to be violated or increase in magnitude, which ensures feasibility when violation is inevitable without sacrificing safety when violation is avoidable.
In rover and tractor-trailer environments we showed that our counterfactual harm constraint enables agents to learn safer and more performant policies than alternative constraint formulations.
The cost of our method is additional computations for counterfactual inference, which was modest in our experiments but could be significant if simulation of the environment model is expensive and counterfactual inference has to be performed over a long time horizon; this could possibly be addressed with off-policy methods.
Finally, our experiments were conducted in single-agent environments with hand-designed default policies.
In future work we will study the value of counterfactual harm constraints in multi-agent settings and explore applications with learned default policies.

\acks{We thank Feng Xiang for the network design and development of the tractor-tractor environment.
We would also like to thank Michal \v{C}\'{a}p, Spencer Sharp, Xiaobai Liu, Yibiao Zhao, and Tobias Gerstenberg for helpful discussions and comments.}

\bibliography{references.bib}
\pagebreak
\appendix
\section{Extensions to Shielding and Hierarchical Methods}\label{app:shield}
This section discusses ways to integrate counterfactual constraints into shielding and hierarchical methods.
We focus on integration of the harm constraint \eqref{eq:harm}, since integrating CCATE will be similar and simpler.
In shielding methods the agent's actions are restricted to be safe either by evaluating an explicit function of the current state and proposed action or implicitly by solving an optimization program.
The shielding can be applied at all steps during training, annealed to be more restrictive as training progresses, or only applied during evaluation/deployment.
In the explicit approach~\citep{hsu2023sim}, we are given a function $d:\mathcal{S}\times\mathcal{A}\rightarrow \R$ and a default controller, $\mu$, that satisfies Assumption \ref{ass:mu}.
$d$ is a discriminator that determines whether or not an action proposed by the learner policy, $a_t^\pi$, is safe.
If it is not, the safe action from $\mu$ is executed.
The shielding controller is implemented as follows:
\begin{align}
    a_t \leftarrow \mu(s_t)\text{ if }d(s_t,a_t^\pi)>0,\ a_t^\pi\text{ otherwise}.
\end{align}
Counterfactual harm can be used as the discriminator by using the SCM \eqref{eq:scm_model} as a predictive model to generate trajectories of the learner policy, applying Algorithm \ref{alg:cf_estimate} to compute the outcome of $\mu$, then estimating \eqref{eq:harm}.
Let 
\begin{align}
    h_g^\mu(s_{t:t+N}) = \ReLU (\hat V^{\pi,\lambda}_g(s_t|s_{t:t+N})-\ReLU(\hat V^{\mu,\lambda}_g(s_t|s_{t:t+N}))),
\end{align}
where $\hat V^{\pi,\lambda}_g(s_t|s_{t:t+N})$ is computed with \eqref{eq:tdl_max} and $\hat V^{\mu,\lambda}_g(s_t|s_{t:t+N})$ with Algorithm \ref{alg:cf_estimate}.
The expected harm given the proposed action $a_t^\pi$ can be used as the discriminator as follows:
\begin{align}
\begin{split}\label{eq:harm_discrim}
    d(s_t,a_t^\pi)=\mathbb{E}&\left[h_g^\mu (s_{t:t+N})\right]\\
\text{where } &s_{t+1}=f(s_t,a_t,\xi_t),\ \xi_t\sim \Xi\\
& a_t=a_t^\pi\\
& a_{t+k}\sim \pi(\cdot |s_{t+k})\text{ for }k\in\{1,...,N-1\}.
\end{split}
\end{align}
\eqref{eq:harm_discrim} can be evaluated using Monte Carlo methods~\citep{james1980monte}.

Implicit shielding approaches~\citep{cheng2019end,shao2021reachability} generate a safe action by solving an optimization program where the discriminator is constrained to be non-positive.
In their most common form, the objective is to minimize deviation from the proposed action:
\begin{subequations}\label{eq:implicit_shield}
    \begin{align}
    a_t^* =\underset{a\in \mathcal{A}}{\text{argmin}}\ &\|a\|\\ \label{eq:implicit_shield_cons}
    \text{s.t. }&d(s_t,a_t^\pi+a)\leq 0,
\end{align}
\end{subequations}
however more elaborate objectives, \emph{e.g.} considering the expected reward, could be considered.
The expected harm \eqref{eq:harm_discrim} can be used as the discriminator in \eqref{eq:implicit_shield_cons}, although this results in \eqref{eq:implicit_shield} becoming a stochastic nonlinear model predictive control (SNMPC) program.
Fortunately, since the decision variable in \eqref{eq:implicit_shield} is only the action for the first timestep, sampling-based methods can be effective if the dimension of $\mathcal{A}$ is low.
Since both of our proposed shielding approaches require additional sampling or solving a SNMPC program, we suggest using shielding only in evaluation/deployment or as a fine-tuning mechanism in the late stages of training.

Hierarchical methods~\citep{barto2003recent,lee2023adaptive} train high and low-level policies $\pi^\text{high}$ and $\pi^\text{low}$ with distinct state/action/reward spaces and functions $(\mathcal{S}^\text{high},\mathcal{A}^\text{high},r^\text{high})$ and $(\mathcal{S}^\text{low},\mathcal{A}^\text{low},r^\text{low})$.
The low-level policy is structured around different behaviors, and the high-level action conditions the low-level policy to either execute a specific behavior or prioritize different objectives.
The following optimization programs are solved~\citep{lee2023adaptive}:
\begin{align}
    \pi^\text{low*}=\underset{\pi^\text{low}}{\text{argmax}}\ \mathbb{E}_{\substack{
    a_t^\text{low}\sim \pi^\text{low}(\cdot|s_t^\text{low},a_t^\text{high})\\
     a_t^\text{high}\sim \pi^\text{high}(\cdot|s_t^\text{high})
    }}\left[\sum_{t\geq 0}\gamma^t r^\text{low}(s_t^\text{low},a_t^\text{low},a_t^\text{high})\right]\\
     \pi^\text{high*}=\underset{\pi^\text{high}}{\text{argmax}}\ \mathbb{E}_{\substack{
    a_t^\text{low}\sim \pi^\text{low}(\cdot|s_t^\text{low},a_t^\text{high})\\
     a_t^\text{high}\sim \pi^\text{high}(\cdot|s_t^\text{high})
    }}\left[\sum_{t\geq 0}\gamma^t r^\text{high}(s_t^\text{high},a_t^\text{low},a_t^\text{high})\right].
\end{align}
If we assume $\mathcal{S}^\text{low}\subseteq \mathcal{S}^\text{high}$, or at the very least features related to constraint satisfaction are included in $\mathcal{S}^\text{high}$, it is straightforward to include \eqref{eq:harm} as a constraint on the high-level policy:
\begin{subequations}
    \begin{align}
        \pi^\text{high*}=\underset{\pi^\text{high}}{\text{argmax}}\ &\mathbb{E}_{\substack{
    a_t^\text{low}\sim \pi^\text{low}(\cdot|s_t^\text{low},a_t^\text{high})\\
     a_t^\text{high}\sim \pi^\text{high}(\cdot|s_t^\text{high})
    }}\left[\sum_{t\geq 0}\gamma^t r^\text{high}(s_t^\text{high},a_t^\text{low},a_t^\text{high})\right]\\
    \text{s.t. }&\mathbb{E}_{\substack{
    a_t^\text{low}\sim \pi^\text{low}(\cdot|s_t^\text{low},a_t^\text{high})\\
     a_t^\text{high}\sim \pi^\text{high}(\cdot|s_t^\text{high})
    }}\left[\max_{t\geq 0} \gamma^t H_g^\mu\left(s_{t}^{\text{high}}\right)\right]=0,
    \end{align}
\end{subequations}
which ensures that the high-level policy does not incur harm over selecting the default policy.
In architectures where the low-level policy is a set of policies (including the default policy) and the high-level action chooses from this set, one may consider further extensions using the simultaneous comparison from~\cite{beckers2022quantifying}, where the constraint violations of each policy are compared to every other policy in the set.
\section{TD($\lambda)$ Estimate for Max Operator}\label{app:max_proof}
This section provides convergence analysis of the operator, $\mathcal{M}$, \begin{align}\label{eq:max_op}
\mathcal{M}V(s_t)=\max\{g(s_t),\gamma(\lambda \mathcal{M}V(s_{t+1})+(1-\lambda)V(s_{t+1}))\}
\end{align}
where $V:\mathcal{S}\rightarrow \R$ is a function approximation of the value function $V(s_t)\approx\mathbb{E}[\max_{\tau \geq t}\gamma^{\tau-t}g(s_\tau)]$.
We show that $\mathcal{M}$ is a contraction and that its fixed point is the true value function.
We first introduce the following Lemma:
\begin{lemma}\label{lem:max_diff}
Given $x,y,z\in \R,\ |\max(x,y)-\max(x,z)|\leq |y-z|$
\end{lemma}
\begin{proof}
The proof is accomplished by exhaustively checking all combinations of $x,\ y$ and $z$.
$x$ lower bounding $y$ and $z$, gives $y-z$.
$x$ upper bounding $y$ and $z$, gives $0$.
$y\geq x\geq z$, gives $y-x$; which is less than $|y-z|$ since $x\geq z$.
$z\geq x\geq y$, gives $z-x$; which is less than $|z-y|$ since $x\geq y$.
\end{proof}
\begin{theorem}
    $\mathcal{M}$ is a contraction with $\gamma \in [0,1)$ and $\lambda \in [0,1)$. Given two value functions $V_1$ and $V_2$, and any state $s\in \mathcal{S}$, $|\mathcal{M}V_1(s)-\mathcal{M}V_2(s)|\leq \eta \|V_1-V_2\|_\infty$, where $\eta \in[0,1)$.
\end{theorem}
\begin{proof}
Writing out the left side of the inequality gives
   \begin{align}\label{eq:step1}\begin{split}
   |\mathcal{M}V_1(s_{t})-\mathcal{M}V_2(s_{t})|= |\max\{ g(s_{t}) ,\gamma (\lambda \mathcal{M}V_1(s_{t+1}) +(1-\lambda)V_1(s_{t+1}))\}-\\-\max\{ g(s_{t}) ,\gamma (\lambda \mathcal{M}V_2(s_{t+1}) +(1-\lambda)V_2(s_{t+1}))\}|.\end{split}
\end{align} 
By Lemma \ref{lem:max_diff} and the triangle inequality
\begin{align}
    \eqref{eq:step1}&\leq |\gamma (\lambda \mathcal{M}V_1(s_{t+1}) +(1-\lambda)V_1(s_{t+1}))-\gamma (\lambda \mathcal{M}V_2(s_{t+1}) +(1-\lambda)V_2(s_{t+1}))|\\
    &  \leq \gamma \lambda |\mathcal{M}V_1(s_{t+1}) -\mathcal{M}V_2(s_{t+1})| + \gamma(1-\lambda)|V_1(s_{t+1})-V_2(s_{t+1})|\label{eq:step2}
\end{align}
Let $\Delta \mathcal{M}V(s)=|\mathcal{M}V_1(s)-\mathcal{M}V_2(s)|$ and  $\Delta V(s)=|V_1(s)-V_2(s)|$.
Repeating the process in (\ref{eq:step1}-\ref{eq:step2}) $n$ times on the first term in \eqref{eq:step2} produces the upper bound:
\begin{align}\label{eq:step3}
  |\mathcal{M}V_1(s_{t})-\mathcal{M}V_2(s_{t})|\leq  \gamma^n\lambda^n\Delta \mathcal{M}V(s_{t+n}) +\sum_{k=1}^n\lambda^{k-1}\gamma^k(1-\lambda) \Delta V(s_{t+k}).
\end{align}
Taking the limit of \eqref{eq:step3} as $n\rightarrow\infty$ with $\gamma\in[0,1]$ and $\lambda\in[0,1)$ gives 
\begin{align}\label{eq:inf_sum}
    |\mathcal{M}V_1(s_{t})-\mathcal{M}V_2(s_{t})|\leq \sum_{k\geq 1}\lambda^{k-1}\gamma^k(1-\lambda) \Delta V(s_{t+k}),
\end{align}
which is the weighted sum of $\Delta V$ at different states.
Proving that the sum of the weights is less than 1 proves $\eta \|V_1-V_2\|_\infty$ upper bounds \eqref{eq:inf_sum}.
For $\gamma \in [0,1)$ and $\lambda \in [0,1)$:
\begin{align}
   \eta= \sum_{k\geq 1}\lambda^{k-1}\gamma^k(1-\lambda) &=\sum_{k\geq 0}\lambda^{k}\gamma^{k+1}(1-\lambda)< \sum_{k\geq 0}\lambda^{k}(1-\lambda)=1
\end{align}
\end{proof}
\begin{theorem}
    The fixed point of $\mathcal{M}$ is the value function $V(s_t)=\mathbb{E}[\max_{\tau \geq t}\gamma^{\tau-t}g(s_\tau)]$
\end{theorem}
\begin{proof}
The proof is accomplished by showing that: $ \mathcal{M}V(s)-V(s) =0$ for any $s\in \mathcal{S}$.
\begin{align}\label{eq:proc2_start}
   \mathcal{M}V(s_t)-V(s_t)= \max\{ g(s_t) ,\gamma (\lambda \mathcal{M}V(s_{t+1}) +(1-\lambda)V(s_{t+1}))\}-V(s_t)\\
    =\max\{ g(s_t) ,\gamma (\lambda \mathcal{M}V(s_{t+1}) +(1-\lambda)V(s_{t+1}))\}-\max\{g(s_t),\gamma V(s_{t+1})\}.
\end{align}
Applying Lemma \ref{lem:max_diff} gives
\begin{align}
   | \mathcal{M}V(s_t)-V(s_t)| \leq& |\gamma (\lambda \mathcal{M}V^\pi(s_{t}) +(1-\lambda)V(s_{t+1})- V(s_{t+1}))|\\\label{eq:proc2_end}
   \leq& \gamma \lambda| \mathcal{M}V(s_{t+1})-V(s_{t+1})|
\end{align}
Let $\Delta \mathcal{M}V(s)= | \mathcal{M}V(s)-V(s)|$
Repeating the process in (\ref{eq:proc2_start}-\ref{eq:proc2_end}) $n$ times gives the following sequence of upper bounds:
\begin{align}
\Delta \mathcal{M}V(s_{t})\leq \gamma \lambda \Delta \mathcal{M}V(s_{t+1})\leq \gamma^2 \lambda^2 \Delta \mathcal{M}V(s_{t+2})\leq ...\leq \gamma^n \lambda^n \Delta \mathcal{M}V(s_{t+n}),
\end{align}
and $\lim_{n\rightarrow\infty}\gamma^n \lambda^n \Delta \mathcal{M}V(s_n)=0$ for $\gamma\in[0,1]$ and $\lambda\in[0,1)$.
\end{proof}
\section{Extended Results}\label{app:results}
This section provides details on hyperparameters, each environment, and training performance.
\subsection{Training Hyperparameters}
The hyperparameters for PPO and the Lagrange multipliers are shown in Table \ref{tab:hyper}.
These were used for all methods.
\begin{table}[h]
    \centering
    \begin{tabular}{|p{0.35\linewidth}|p{0.40\linewidth}|}
     \hline Parameter   &  Value \\
      \hline Optimizer   &  Adam with learning rate 1e-3 \\
      \hline Max gradient & 1.0\\
      \hline Discount factor, $\gamma$ & 0.99\\
      \hline Stability parameter, $\lambda$ & 0.95\\
      \hline Entropy coefficient & 0.01\\
      \hline Clipping parameter & 0.2\\
      \hline Steps between updates, $M$ & 24\\
      \hline Minibatches per update & 3\\
      \hline Epochs per update & 5\\
       \hline Lagrange multiplier learning rate (rover) & 1e-3 for 250 updates, linearly increase to 1 at 15,000 updates\\
      \hline Lagrange multiplier learning rate (tractor-trailer) & 1e-3 for 1,000 updates, linearly increase to 1e-1 at 15,000 updates\\
      \hline
    \end{tabular}
    \caption{Training Hyperparameters}
    \label{tab:hyper}
\end{table}
\subsection{Rover}
The rover is modeled as a kinematic bicycle model with a wheelbase of 0.5 m and a circular footprint with of radius 0.5 m.
The action commands are an acceleration limited to $\pm$ 1 m/s$^2$ and wheel angle limited to 0.5 rad.
The speed is limited to $\pm$ 1 m/s.
The road surface varies with a friction coefficient of $\rho\in[0.3,1.0]$ that is drawn from a uniform distribution for each environment.
Friction affects the dynamics model by clipping the yawrate and acceleration to lie within a circle where $\rho=1$ allows maximum longitudinal and lateral acceleration.
The rover receives a noisy observation of its state and the friction coefficient.
Normally distributed noise is also added to the rover's action commands and the true friction coefficient at each timestep.
The rover reaches the goal if it arrives within 0.5 m of the origin with a speed $\leq$ 0.1 m/s.
We use a timestep of 0.5 s and episode length of 100 steps.
The rover uses 3,000 parallel environments.

\begin{figure}
    \centering
    \includegraphics[width=0.9\textwidth]{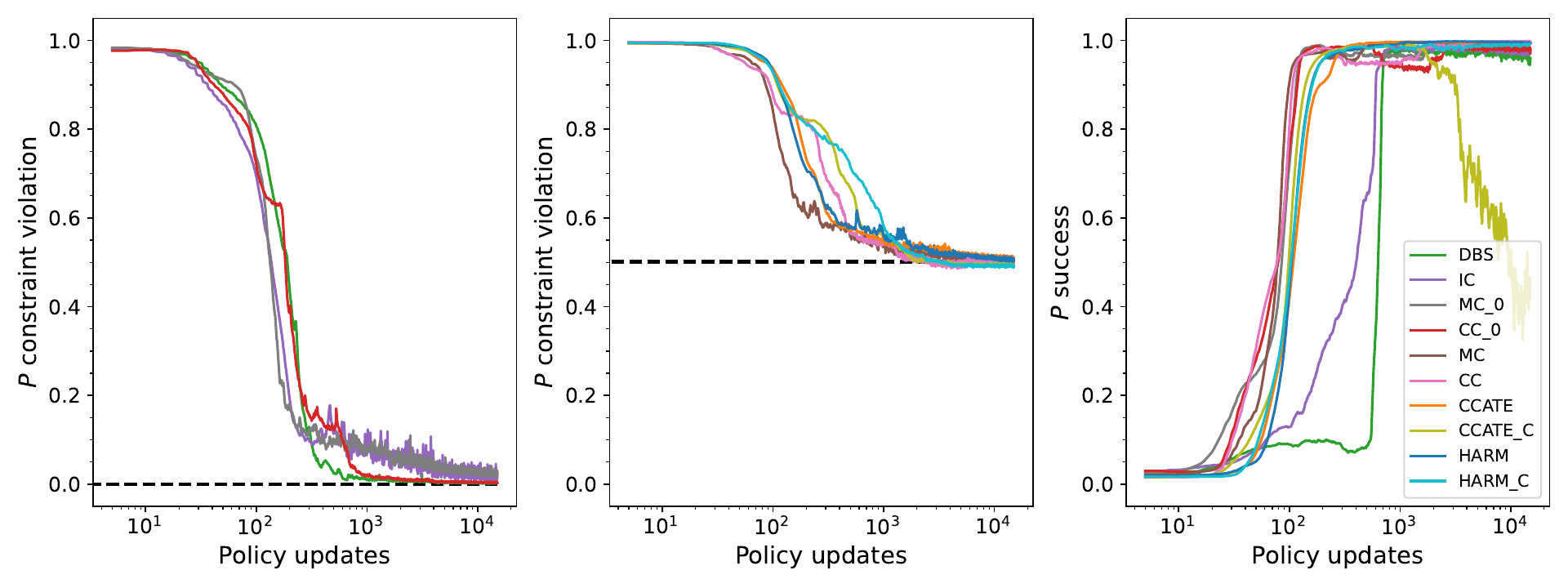}
    \caption{Performance during training of rover environment. From left to right: probability of constraint violation for methods with the feasible initialization (DBS, IC, MC\_0, CC\_0); probability of constraint violation for methods initialized outside of the viability kernel (MC, CC, CCATE, CCATE\_C, HARM, HARM\_C); success rate.
    The black dashed line indicates the probability of constraint violation for executing the default policy, $\mu$, from the initial state distribution.
    }
    \label{fig:rover_training}
\end{figure}

Figure \ref{fig:rover_training} shows the probability of constraint violation and success rate for each algorithm during training.
CCATE\_C becomes unstable as training progresses and its performance collapses.
We believe this is because \eqref{eq:ccate} uses the estimated value function of the default policy as a threshold at each step, which can be biased and lead to unfair penalization as the Lagrange multipliers increase. 
HARM discounts this bias by $\sim N$ steps by using the estimator \eqref{eq:tdl_max} in Algorithm \ref{alg:cf_estimate}.
\subsection{Tractor-Trailer Parking}
For the tractor-trailer, the actions are the acceleration and wheel angle rate and are limited to magnitudes of 2 m/s$^2$ and 1 rad/s respectively.
The speed is limited to $v\in$[-2,5] m/s.
Normally distributed noise proportional to the speed is added to the joint angle, yawrate, and lateral velocity of each agent.
Additionally the trailers take uncertain wheelbases sampled from a normal distribution.
The static environment is an occupancy grid and the agent performs raycasting with 32 rays that return distance to the closest occupied points and the relative velocity to the agent.
Uniformly distributed noise proportional to the distance of the obstacle point is added to the observation.
We use a timestep of 0.5 s and episode length of 300 steps. The tractor trailer uses 10,000 parallel environments.

\begin{figure}
    \centering
    \includegraphics[width=0.9\textwidth]{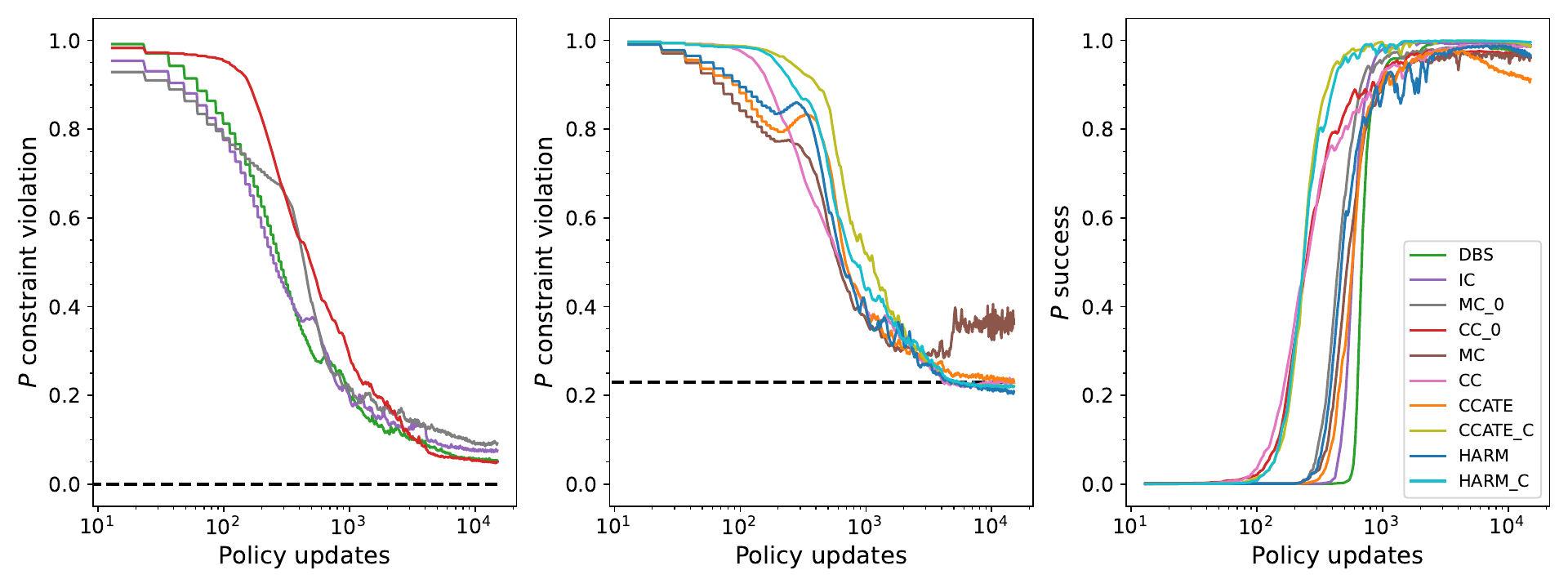}
    \caption{Performance during training of tractor trailer environment.
    From left to right: probability of constraint violation for methods with the feasible initialization (DBS, IC, MC\_0, CC\_0); probability of constraint violation for methods initialized outside of the viability kernel (MC, CC, CCATE, CCATE\_C, HARM, HARM\_C); success rate.
    The black dashed line indicates the probability of constraint violation for executing the default policy, $\mu$, from the initial state distribution.
    }
    \label{fig:tractor_trailer_training}
\end{figure}

Figure \ref{fig:tractor_trailer_training} shows the probability of constraint violation and success rate for each algorithm during training.
Compared to the rover environment, methods with the feasible initialization (DBS, IC, MC\_0, CC\_0) have a higher (5-10\%) probability of constraint violation after 15,000 policy updates.
We attribute the difference to the trailer parking task being more complex and the fact that the raycasting observation model may not fully capture features of the occupancy grid.
It is possible that training for longer would eventually decrease the constraint violations.
Additionally, using a recurrent network to capture the observation history could improve performance.
Of the policies initialized outside of the viability kernel, MC is the least safe; and we even see an increase in the probability of constraint violations in the middle of training.
As discussed in Section \ref{sec:formulation}, MC uses \eqref{eq:initial_cons} which constrains the expected magnitude of constraint violation over the initial state distribution, $p_0$.
This does not result in a safe policy because the learner can reduce large constraint violations while incurring small ones, and the constraint will be satisfied as long as the average is low enough.
CC, CCATE, CCATE\_C, HARM, and HARM\_C all either beat or match the probability of violation obtained by sampling the default policy, with HARM achieving the lowest.

\end{document}